\documentclass{aamas2015}
\usepackage{hyperref}
\usepackage{url}
\usepackage{algorithm}
\usepackage{algorithmic}
\usepackage{float}
\usepackage{graphicx}
\usepackage{epstopdf}
\usepackage{stfloats}
\usepackage{amsfonts}
\usepackage{cite}

\begin{document}


\title{DJ-MC: A Reinforcement-Learning Agent for Music Playlist Recommendation}

\numberofauthors{3}
\author{
\alignauthor 
Elad Liebman \\ \affaddr{Computer Science Department}\\
			\affaddr{The University of Texas at Austin}			\\
			\affaddr{eladlieb@cs.utexas.edu}
\alignauthor
Maytal Saar-Tsechansky \\ \affaddr{McCombs School of Business}\\
			\affaddr{The University of Texas at Austin}			\\
			\affaddr{maytal@mail.utexas.edu}			
\alignauthor 
Peter Stone \\ \affaddr{Computer Science Department}\\
			\affaddr{The University of Texas at Austin}\\
			\affaddr{pstone@cs.utexas.edu}
}

\maketitle

\begin{abstract}
In recent years, there has been growing focus on the study of automated recommender systems. Music recommendation systems serve as a prominent domain for such works, both from an academic and a commercial perspective. A fundamental aspect of music perception is that music is experienced in temporal context and in sequence. In this work we present DJ-MC, a novel reinforcement-learning framework for music recommendation that does not recommend songs individually but rather song sequences, or playlists, based on a model of preferences for both songs and song transitions. The model is learned online and is uniquely adapted for each listener. To reduce exploration time, DJ-MC exploits user feedback to initialize a model, which it subsequently updates by reinforcement. We evaluate our framework with human participants using both real song and playlist data. Our results indicate that DJ-MC's ability to recommend sequences of songs provides a significant improvement over more straightforward approaches, which do not take transitions into account.
\end{abstract}

\section{Introduction}

Music is one of the most widespread and prevalent expressions of human culture. It has accompanied the human experience throughout history, and the enjoyment of music is one of the most common human activities. As an activity, music listening sessions commonly span over a sequence of songs, rather than a single song in isolation. Importantly, it is well established that music is experienced in temporal context and in sequence \cite{davies1978psychology,kahnx1997patterns}. This phenomenon not only underlies the notion of structure in music (as in the canonical sonata form \cite{cook1994guide}), but also implies that the pleasure one derives from a complete song is directly affected by its relative position in a sequence. This notion also underlies the manner in which DJs construct  playlists \cite{cliff2000hang}, and indeed,  research on automated playlist construction has aimed to produce generally appealing playlists \cite{oliver2006papa,crampes2007automatic}. However, such works have not considered the construction of personalized playlists tailored to \emph{individual} users' preferences. 

In the field of recommender systems, \cite{adomavicius2005toward} music has been of particular interest, both academically \cite{adomavicius2005toward,o2005trust} and commercially \cite{barrington2009smarter}. Pandora, Jango, and Last.fm are some examples of  popular contemporary commercial applications. To the best of our knowledge, however, research on \emph{personalized} music recommendations has focused mostly on predicting users' preferences over \emph{individual} songs, rather than song \emph{sequences}. 
 
Overall, there has been little effort to relate learning individual
listener preferences with holistic playlist generation. In this paper,
we aim to bridge this gap and present DJ-MC, a novel framework for
\emph{adaptive, personalized} music playlist recommendation. In this
framework, we formulate the playlist recommendation problem as a
sequential decision making task, and borrow tools from the
reinforcement learning literature to learn preferences over both songs
and song transitions on the fly. Our contributions are as
follows. First, we formulate the problem of selecting which sequence
of songs to play as a Markov Decision Process, and demonstrate the
potential effectiveness of a reinforcement-learning based approach in
a new practical domain. Second, we test the hypothesis that sequence
does have a significant effect on listener experience through a user
study. Third, we show empirically that DJ-MC's account for song order
allows it to outperform recommendations based strictly on individual
song preferences, implying such preferences can be learned efficiently
with limited user information. In particular, we demonstrate that
starting with no knowledge of a new user's preferences, DJ-MC is able
to generate personalized song sequences within a single listening
session of just 25--50 songs.

The remainder of this paper is organized as follows. In Section $2$ we discuss our reformulation of playlist generation as a reinforcement learning task. In Section $3$ we describe how the DJ-MC agent models different aspects of the MDP for the purpose of learning. In Section $4$ we present the real-world data sources we used in this paper. In Section $5$ we present the full DJ-MC agent architecture. In Section $6$ we discuss the performance of DJ-MC in simulation, and in Section $7$ we present the results of applying DJ-MC in a user study with human participants. In Section $8$ we discuss related work and put our contributions in a broader context, and finally in Section $9$ we summarize and discuss our results.

\section{Reinforcement Learning Framework}

We consider the adaptive playlist generation problem formally as an
episodic Markov Decision Process (MDP). An episodic MDP is a tuple
$(S, A, P, R, T)$ where $S$ is the set of states; $A$ the set of
actions, $P:S \times A \times S \rightarrow [0,1]$ is the state transition probability function where $P(s,a,s')=r$ denotes the probability of transitioning from state $s$ to state $s'$ when taking action $a$. $R:S \times A \rightarrow \mathbb{R}$ is the state-action reward function, where $R(s,a)=r$ means that taking action $a$ from state $s$ will yield reward $r$. $T$ is the set of terminal states, which end the episode. 

For the purposes of our specific application, consider a finite set of $n$ musical tracks (songs)
  $\cal{M}$ $= \{a_1, a_2, \ldots, a_n \} $
   and assume that playlists are of length
  $k$.  Our MDP formulation of the music playlist recommendation
task is then as follows.

\begin{itemize}
\item To capture the complex dependency of listener experience on the
  entire sequence of songs heard, a Markov state must include an
  ordered list of all prior songs in the playlist.  Thus, the state space $S$ is the entire ordered sequence of songs played, $S = \{(a_1, a_2,
  \ldots, a_i) | 1 \leq i \leq k; \forall j \leq i, a_j \in
  \cal{M}\}$.

  That is, a state $s \in S$ is an ordered tuple of songs ranging in
  length from 0 when choosing the first song of the playlist to $k$
  when the playlist is complete.

\item The set of actions $A$ is the selection of the next song to
play, $a_k \in A$. This means that the action space is exactly the set of songs:
$A = \cal{M}$.

\item These
definitions of $S$ and $A$ induce a deterministic transition function
$P$. As such, we can use the shorthand notation $P(s,a) = s'$ to
indicate that when taking action $a$ in state $s$, the probability of
transitioning to $s'$ is 1, and to $s'' \neq s'$ is 0.  Specifically, \linebreak
$P((a_1, a_2, \ldots, a_i), a^*) = (a_1, a_2, \ldots, a_i, a^*)$.

\item $R(s,a)$ is the utility (or pleasure) the current listener
derives from hearing song $a$ when in state $s$.  Note that this
formulation implies that each listener induces a unique reward
function.  A key challenge addressed in this paper is enabling
efficient learning of $R$ for a new listener.

\item $T = \{(a_1, a_2, \ldots m_k)\}$: the set of playlists of length
$k$.
\end{itemize}

\emph{Solving} an MDP typically refers to finding a policy $\pi: S
\rightarrow A$ such that from any given state $s$, executing action
$\pi(s)$ and then acting optimally (following the optimal policy
$\pi^*$) thereafter, yields the highest (expected) sum of rewards over
the length of the episode.  In our case, since $P$ is deterministic,
$\pi*$ corresponds to the single sequence of songs that would be most
pleasing to the listener.\footnote{We consider the problem as finding
a single playlist in isolation, ignoring the fact that the same
listener may not want to hear similar sequences repeatedly.  In
practice, the stochasticity of our approach makes it exceedingly
unlikely that the same sequence would be presented to a given listener
multiple times, as will become clear below.}  However, we assume that
the listener's reward function $R$ is initially unknown.  We consider
the fundamental challenge of playlist generation as being efficiently
modeling $R$.

In particular, in the reinforcement learning literature, there are two
high-level approaches to approximating (learning) $\pi^*$: model-free
and model-based.  \emph{Model-free} approaches learn the value of
taking an action $a$ from state $s$ directly.  Typical approaches,
such as $Q$-learning and SARSA \cite{Sutton1998} are computationally
efficient and elegant, but require a lot of experiential data to
learn.  \emph{Model-based} approaches alternatively learn the
transition and reward functions ($P$ and $R$) so as to be able to
\emph{simulate} arbitrary amounts of experiential data in order to
find an approximate solution to the MDP in an approach that can be
thought of as \emph{planning} through forward lookahead search.
Compared to model-free methods, most model-based algorithms are
significantly more computationally expensive, especially if they
re-solve the MDP whenever the model changes.  However, in many
applications, including playlist recommendation, where data is
considerably more scarce than computation, this tradeoff of
computational expense for data efficiency is a good one.  We therefore
adopt a model-based learning approach in this paper (see
Sections \ref{sec:model} and \ref{sec:djmc} for details).

In the MDP defined above, the transition function $P$ is trivially
known.  Therefore the only unknown element of the model necessary for
model-based learning is $R$, the current listener's utility
(enjoyment) function.  Indeed modeling $R$ in such a way that
generalizes aggressively and accurately across both songs and song
transitions is the biggest technical challenge in this work.  Consider
that even for a moderately sized music corpus of $10^3$ songs, and for
playlist horizons of $10$ songs, the size of the state space alone
would be $10^{30}$. It is impractical for a learning agent to even
explore any appreciable size of this state space, let alone learn the
listener's utility for each possible state (indeed our objective is to
learn a new user's preferences and generate a personalized song
sequence within a single listening session of 25--50 songs).
Therefore to learn efficiently, the agent must internally represent
states and actions in such a way that enables generalization of the
listener's preferences.

Section $3$ presents how DJ-MC compactly represents $R$ by 1)
generalizing across songs via a factored representation; and 2)
separating $R$ into two distinct components, one dependent only on the
current song ($a$), and one dependent on the transition from the past
history of songs to the current song ($s$ to $a$).  Recognizing that
DJ-MC's specific representation of $R$ is just one of many possible
options, we also evaluate the extent to which the representational
choices made are effective for generalization and learning.

\section{Modeling}
\label{sec:model}
As motivated in the previous section, learning a listener's preference
function over a large set of songs and sequences requires a compact
representation of songs that is still rich enough to capture
meaningful differences in how they are perceived by listeners.  To
this end, we represent each song as a vector of song \emph{descriptors}.  

Specifically DJ-MC uses spectral auditory descriptors that include
details about the spectral fingerprint of the song, its rhythmic
characteristics, its overall loudness, and their change over time. We
find that these descriptors enable a great deal of flexibility (for
instance, in capturing similarities between songs from vastly
different backgrounds, or the ability to model songs in unknown
languages). Nonetheless, our framework is in principle robust to using
any sufficiently expressive vector of song descriptors. Section $3.1$
specifies in detail the descriptors used by DJ-MC.


In order to further speed up learning, we make a second key
representational choice, namely that the reward function $R$
corresponding to a listener can be factored as the sum of two distinct
components: 1) the listener's preference over \emph{songs} in
isolation, $R_s: A \rightarrow \mathbb{R}$ and 2) his preference over
\emph{transitions} from past songs to a new song, $R_t: S \times A
\rightarrow \mathbb{R}$.  That is, $R(s,a) = R_s(a) + R_t(s,a)$.

Section $3.2$ describes DJ-MC's reward model in detail.  Section $3.3$
then evaluates the extent to which the chosen descriptors are able to
differentiate meaningfully between song sequences that are clearly
good and clearly bad.  


\subsection{Modeling Songs}

As motivated above, we assume each song can be factored as a vector of
scalar descriptors that reflect details about the spectral fingerprint
of the song, its rhythmic characteristics, its overall loudness, and
their change over time. For the purpose of our experiments, we used
the acoustic features in the Million Song Dataset representation
\cite{bertin2011million} to extract $12$ meta-descriptors, out of
which $2$ are $12$-dimensional, resulting in a $34$-dimensional song
descriptor vector. The complete set of descriptors is summarized in
Table $1$.


\begin{table}
\begin{center}

    \begin{tabular}{|l|l|}
    \hline
    Descriptors                           & Descriptor Indices \\ \hline
    10th and 90th percentiles of tempo    & 1,2             \\
    average and variance of tempo         & 3,4             \\
    10th and 90th percentiles of loudness & 5,6             \\ 
    average and variance of loudness      & 7,8             \\ 
    pitch dominance                       & 9--20            \\
    variance of pitch dominance           & 21              \\
    average timbre weights                & 22--33           \\ 
    variance in timbre                    & 34              \\ \hline
    \end{tabular}
    
    \end{center}
    \caption{{\scriptsize Descriptors used for song representation. Tempo
        data was based on beat durations. Thus the first descriptor is
        the 10th percentile of beat durations in the song.  Loudness
        was straightforwardly obtained from amplitude. Pitch dominance
        weights each of the $12$ possible pitch classes based on their
        presence in the song averaged over time. Timbre weights are
        the average weights of the $12$ basis functions used by the
        Echo Nest analysis tool to capture the spectro-temporal
        landscape of the song.}}
\end{table}

\subsection{Modeling The Listener Reward Function}

Despite an abundance of literature on the psychology of human musical
perception \cite{tan2010psychology}, there is no canonical model of
the human listening experience. In this work we model listening as
being dependent not only on preferences over the descriptors laid out
above, but also over feature \emph{transitions}. This model is fairly
consistent with many observed properties of human perception, such as the stochastic dependence on remembering earlier events, and evidence of working memory having greater emphasis on the present
\cite{davies1978psychology,berz1995working,tan2010psychology}.


We now proceed to specify the two components of $R$: $R_s$ and $R_t$.

\subsubsection{Listener Reward Function over Songs $R_s$}

To model $R_s$, we use a sparse encoding of the song descriptors to
generate a binary feature vector.  $R_s$ is then a linear function of
this feature vector: that is, we assume that each feature contributes
independently to the listener's utility for the song.

Specifically, for each song descriptor, we collect statistics over the
entire music database, and quantize the descriptor into 10-percentile
bins.  
Following standard reinforcement learning notation, we denote the
feature vector for song $a$ as $\theta_s(a)$.  It is a vector of size $\#bins
\times \#descriptors = 10 \times 34 = 340$ consisting of
$\#descriptors$ containing 1's at coordinates that correspond to the bins song $a$ populates, and $0$ otherwise, meaning $\theta_s(a)$ behaves as an indicator function (the weight of $\theta_s(a)$ will be 34 overall).

For each feature, we assume the listener has a value representing the
pleasure they obtains from songs with that feature active.  These values
are represented as a weight vector $\phi_s(u)$.  Thus $R_s(a) =
\phi_s(u) \cdot \theta_s(a)$.  The parameters of $\phi_s(u)$ must be
learned afresh for each new user.

\subsubsection{Listener Reward Function over Transitions $R_t$}

A main premise of this work is that in addition to the actual songs
played, a listener's enjoyment depends on the \emph{sequence} in which
they are played.  To capture this dependence, we assume that
$$E[R_t((a_1, \ldots, a_{t-1}), a_t)] = \sum\limits_{i=1}^{t-1}
\frac{1}{i^2}r_t(a_{t-i},a_t)$$ where $r_t(a_i,a_j)$ represents the
listener's utility for hearing song $a_j$ sometime after having heard
$a_i$.
The term $\frac{1}{i^2}$ represents the notion that a song that was
played $i$ songs in the past has a probability of $\frac{1}{i}$ of
affecting the transition reward (i.e.\ being ``remembered''), and when
it does, its impact decays by a second factor of $\frac{1}{i}$ (its
impact decays over time).

It remains only to specify the song to song transition reward function
$R_t(a_i,a_j)$.  Like $R_s$, we can describe $R_t$ as a linear
function of a sparse binary feature vector: $R_t(a_i,a_j) = \phi_t(u)
\cdot \theta_t(a_i,a_j)$ where $\phi_t(u)$ is a user-dependent weight
vector and $\theta_t$ is a binary feature vector.

Were we to consider the transitions between all 340 features of both
$a_i$ and $a_j$, $\theta_t$ would need to be of length $340^2 >
100,000$.  For the sake of learnability, we limit $\theta_t$ and $\phi_t$ to only
represent transitions between 10-percentile bins of the same song
descriptors.  That is, there is for each of the 34 song descriptors,
there are 100 features, one of which is 1 and 99 of which are 0,
indicating which pair of 10-percentile bins were present in songs
$a_i$ and $a_j$.  Therefore, overall, $\theta_t$ consists of 3,400
binary features, 34 of which are 1's.

Clearly, this representation is limiting in that it cannot capture the
joint dependence of listener utility on transitions between multiple song
descriptors.  Especially for the pitch class features, these are
likely to be relevant.  We make this tradeoff in the interest of
enabling learning from relatively few examples.  Empirical results
indicate that this representation captures enough of real peoples'
transition reward to make a difference in song recommendation quality.

Like $\phi_s(u)$, the parameters of $\phi_t(u)$ must be learned afresh
for each new user.  Thus all in all, there are 3740 weight parameters to
learn for each listener.

With even that many parameters, it is infeasible to experience songs
and transitions with all of them active in just 25 songs.  However
DJ-MC is able to leverage knowledge of even a few transition examples to
plan a future sequence of songs that is biased in favor of the
positive ones and against the negative ones.



\subsection{Expressiveness of the Listener Model}

This representation of the listener's utility function as a
3740-dimensional sparse binary feature vector is just one of many
possible representations. A necessary property of a useful
representation is that its features are able to differentiate between
commonly perceived ``good'' vs. ``bad'' sequences, and the DJ-MC agent
internally relies on this property when modeling the listener reward
function. To evaluate whether our features are expressive enough to
allow this differentiation, we examine the transition profile for two
types of transitions, ``poor'' vs. ``fair'', both derived from the
same population of songs. We generate ``fair'' transitions by sampling
pairs of songs that appeared in an actual sequence. We generate
``poor'' transitions by interleaving songs so that each one is
distinctly different in character (for instance, a fast, loud track
followed by a soft piece). The difference between the two profiles can
be seen in Figure \emph{1}. More definitive evidence in favor of the
adequacy of our representation is provided by the successful empirical
application of our framework, discussed in Section $7$.

\begin{figure}[h!]
  \label{fig1}
  \centering
    \includegraphics[height=150pt,width=0.5\textwidth,natwidth=310,natheight=322]{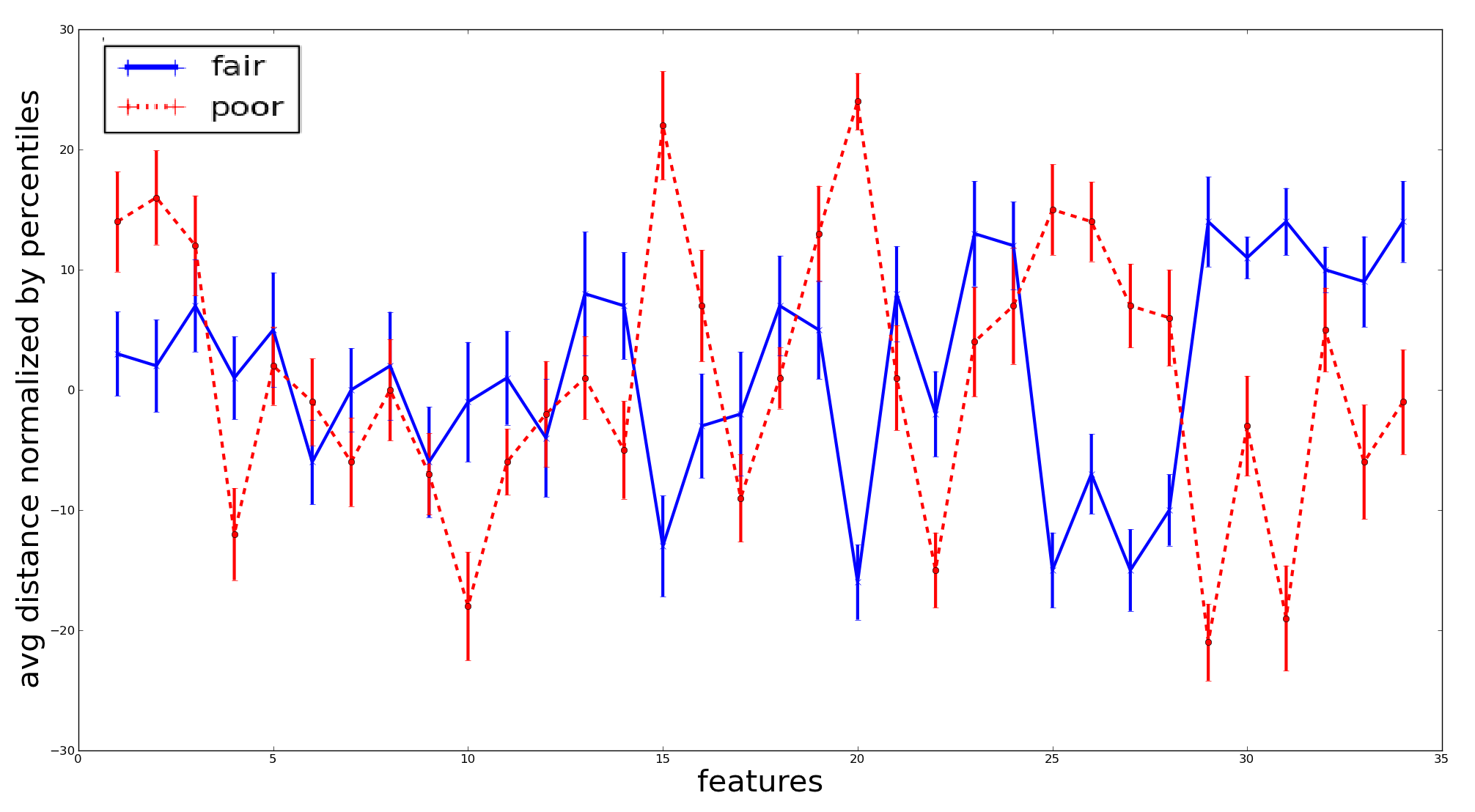}
      \caption{{\scriptsize Example of fair vs. poor transition profiles, based on the same set of $20$ songs. The plot shows the average transition delta for each feature. Both the fair transitions and the poor ones are constructed from the same core set of $20$ songs taken from $5$ different albums. In the case of fair transitions, we maintain the original order. In the case of poor transitions, the albums are randomly interleaved. The results indicate that qualitatively different sequences are indeed distinguishable in our feature model. In this specific example, $19$ of the $34$ features are discriminative (confidence intervals do not overlap). We expect different features to be discriminative for different transition profiles.
}}
\end{figure}

\section{Data}

A significant component of this work involves extracting real-world data for both songs and playlists to rigorously test our approach. In this section we discuss the different data sources we used to model both songs and playlists. For songs, we relied on the Million Song Dataset \cite{bertin2011million}, a freely-available collection of audio features and metadata for a million contemporary popular music tracks. The dataset covers $44,745$ different artists and $10^6$ different tracks. All the features described in Table \emph{1} are derived from this representation. An example of the audio input for a single track is provided in Figure \emph{2}. It should be noted that our agent architecture (described in detail in Section $5$) is agnostic to the choice of a specific song corpus, and we could have easily used a different song archive.

\begin{figure}[h!]
  \label{fig1}
  \centering
    \includegraphics[height=140pt,width=0.5\textwidth,natwidth=310,natheight=322]{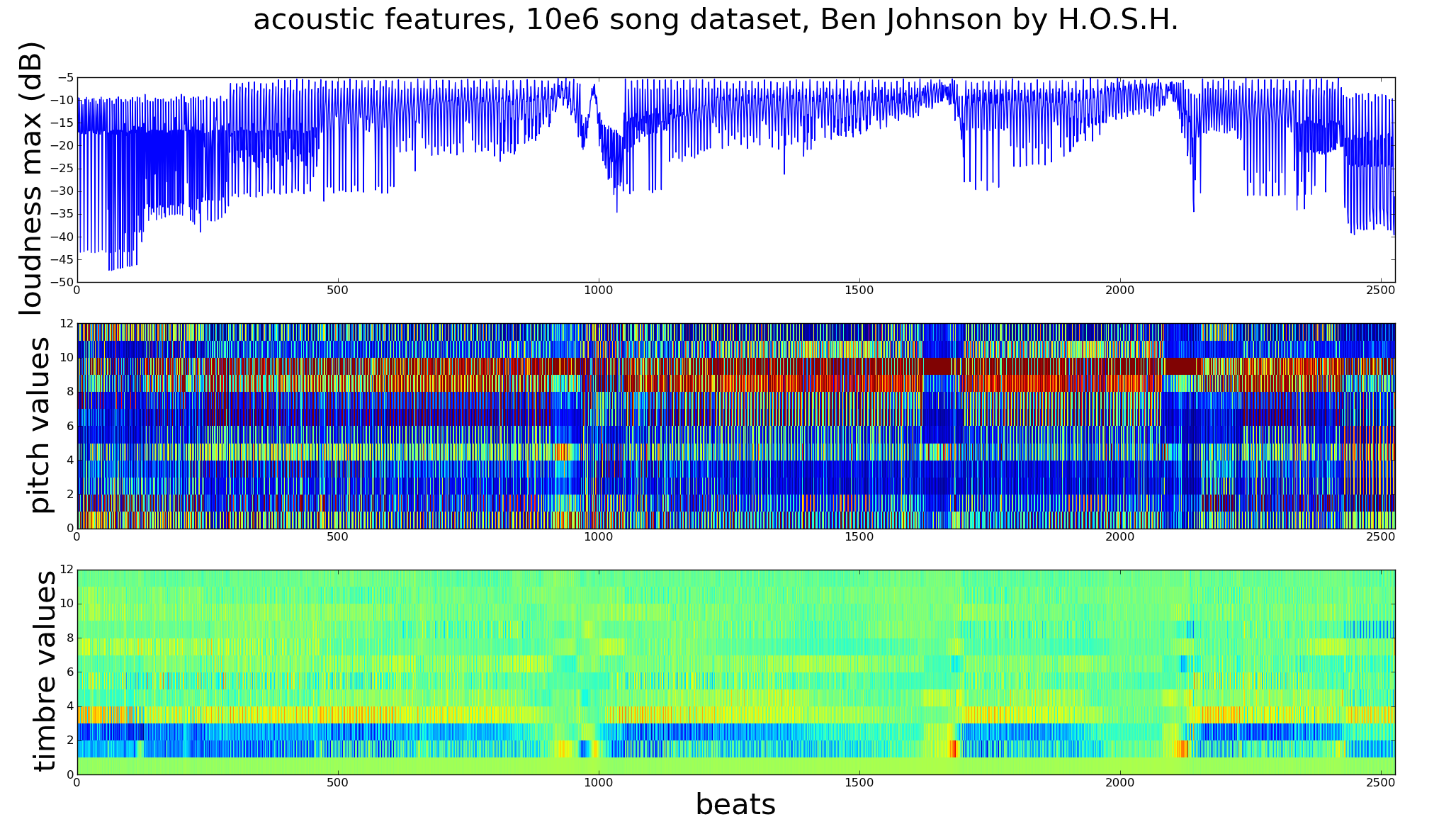}
      \caption{{\scriptsize Example of loudness, pitch and timbre data for an example track over time (time units are in beats). Loudness is in dB, pitch measurements are represented as a vector of $12$ values between $0$ and $1$ representing dominance per pitch class. Timbre is approximated as the decomposition of the spectro-temporal surface of the song according to 12 pre-computed basis functions.
}}
      \end{figure}

To initially test our approach in simulation (a process described in detail in Section $6$), we also needed real playlists to extract song transition data from. A good source of playlists needs to be sufficiently rich and diverse, but also reflect real playlists ``in the wild''. In this paper, we used two separate sources. The first, the Yes.com archive, is corpus collected by Chen et al. \cite{chen2012playlist}. These playlists and related tag data were respectively crawled from Yes.com and Last.fm. Chen et al. harvested data between December 2010 and May 2011, yielding 75,262 songs and 2,840,553 transitions. The second source is the Art of the Mix Archive, collected by Berenzweig et al \cite{berenzweig2004large}. Berenzweig et al. gathered ~29,000 playlists from The Art of the Mix (\url{www.artofthemix.org}), a repository and community center for playlist hobbyists. These playlists were (ostensibly) generated by real individual users, rather than a commercial radio DJ or a recommendation system, making this corpus particularly appealing for listener modeling.

\section{DJ-MC}
\label{sec:djmc}
In this section we introduce DJ-MC, a novel reinforcement learning approach to a playlist-oriented, personalized music recommendation system. The DJ-MC agent architecture contains two major components: learning of the listener parameters ($\phi_s$ and $\phi_t$) and planning a sequence of songs. The learning part is in itself divided into two parts - initialization and learning on the fly. Initialization is critical if we wish to engage listeners quickly without losing their interest before the agent has converged on a good enough model. Learning on the fly enables the system to continually improve until it converges on a reliable model for that listening session. In simulation, we assume the user is able to specify an initial list of songs that they like (this is similar to most initialization practices used by commercial music recommendation systems). However, in Section $7$ we show this step can be replaced with random exploration, while still reaching compelling results at the exploitation stage.

The planning step enables the selection of the next appropriate song to play. As pointed out in Section $2$, given the sheer scope of the learning problem, even after various abstraction steps, solving the MDP exactly is intractable. For this reason we must approximate the solution. From a practical perspective, from any given state, the objective is to find a song that is ``good enough'' to play next. For this purpose we utilize Monte Carlo Tree Search.

In Sections $5.1$ and $5.2$ we describe the initialization steps taken by DJ-MC. In Section $5.3$ we describe the core of the learning algorithm, which learns on the fly. in Section $5.4$ we describe the planning step. The full agent pseudocode is provided in Algorithm 5.

\subsection{Learning Initial Song Preferences}

To initialize the listener's song model, DJ-MC polls the listener for
his $k_s$ favorite songs in the database and passes them as input to
Algorithm 1.  As a form of smoothing (or of maintaining a uniform
prior), each element of $\phi_s(u)$ is initialized to $1/(k_s+\#bins)$,
where $\#bins$ is the granularity of discretization of each song
descriptor -- in our case 10 (line 2).  Then for each favorite song
$a$, $\phi_s(u)$ is incremented by $1/(k+\#bins) \cdot \theta_s(a)$
(line 5).  At the end of this process, the weights in $\phi_s(u)$
corresponding to each song descriptor sum to 1.



\begin{algorithm}[tb!]
   \caption{Initialize Song Preferences $R_s$}
\label{alg:oneshot}
\begin{algorithmic}[1]
\STATE {\bfseries Input:} Song corpus, $\cal{M}$
 \vskip 1pt Number of preferred songs to be provided by listener, $k_s$
 \vskip 1pt
\STATE initialize  all coordinates of $\phi_s$ to $1/(k_s+\#bins)$
\STATE \emph{preferredSet} = $\{a_1, \ldots, a_{k_s} \}$ \emph{(chosen by the listener)}
   \FOR{$i=1$ {\bfseries to} $k_s$}
        \STATE {$\phi_s = \phi_s + \frac{1}{(k_s+1)} \cdot \theta_s(a_i)$}
\ENDFOR

\end{algorithmic}

\end{algorithm}

\subsection{Learning Initial Transition Preferences}


In the second stage, the listener is queried for preferences regarding transitions, following the procedure in Algorithm 2. As in the case of initializing song preferences, the predicted value of a transition from bin $i$ to bin $j$ for each feature is initialized to $1/(k_t+\#bins)$ where $k_t$ is the number of transitions queried and $\#bins$ is the number of feature transition bins -- in our case 100 (line 2).

We wouldn't want to query transitions for too small a subset of preferred songs, because that won't necessarily reveal enough about the preferred transitions. For this reason we explore the preferences of the listener in a targeted fashion, by presenting them with different possible transitions that encapsulate the variety in the dataset, and directly asking which of a possible set of options the listener would prefer. On the other hand, we would also like to exclude regions in the search space where expected song rewards are low. 

To accomplish both ends, DJ-MC first chooses a $50$-\% subset of the songs $\cal{M}^*$ of the song corpus $\cal{M}$ which, based on its song rewards model, obtains the highest song reward $R_s$ (line 3). Then, DJ-MC queries transition preferences over this upper median of songs by eliciting user feedback. It does so by applying the $\delta$-medoids algorithm, a novel method for representative selection (line 5) \cite{repsel}. This algorithm returns a compact but close-fitting subset of representatives such that no sample in the dataset is more than a parameter $\delta$ away from a representative, thus providing a diverse sample of the upper median of songs. $\delta$ is initialized to be the $10$-th percentile of the distance histogram between all pairs of songs in the database (line 4). We denote the representative subset $\cal{C}$. To model transitions, DJ-MC chooses songs from $\cal{C}$, and queries the listener which song $a_i \in \cal{C}$ they would like to listen to next (line 8).\footnote{If $\cal{C}$ is too large, it can be replaced at this step with a smaller subset, depending on the parameters of the system and the size of $\cal{C}$.}  For modeling purposes, we assume the listener chooses the next song he would prefer by simulating the listening experience, including the non-deterministic history-dependent transition reward, and choosing the one with the maximal total reward. DJ-MC then proceeds to update the characteristics of this transition, by increasing the weight of transition features by $1/(k+\#bins)$ (line 9) , similarly to how it updated the model for song preferences (so again, the weights of each individual descriptor sum up to $1$). The full details of the algorithm are described in Algorithm 2.

\begin{algorithm}[tb!]
   \caption{Initialize Transition Preferences $R_t$}
\label{alg:oneshot}
\begin{algorithmic}[1]
\STATE {\bfseries Input:} Song corpus $\cal{M}$ 
 \vskip 1pt Number of transitions to poll the listener, $k_t$
 \vskip 1pt
\STATE initialize  all coordinates of $\phi_t$ to $1/(k_t+\#bins)$
\STATE Select upper median of $\cal{M}$, $\cal{M}^*$, based on $R_s$
\STATE $\delta = $ 10th percentile of all pairwise distances between songs in $\cal{M}$
\STATE representative set $ \cal{C} = \delta$ -medoids $(\cal{M}^*)$
\STATE $song_0 = $ choose a song randomly from $\cal{C}$
\FOR {$i=1$ {\bfseries to} $k_t$}
\STATE $song_i \leftarrow  $ \emph{ chosen by the listener from $\cal{C}$}
\STATE $\phi_t = \phi_t + \frac{1}{(k_t+1)} \cdot \theta_t(song_{i-1}, song_i)$
\ENDFOR
\end{algorithmic}

\end{algorithm}

\subsection{Learning on the fly}

After initialization, DJ-MC begins playing songs for the listener,
requesting feedback, and updating $\phi_s$ and $\phi_t$ accordingly.
For ease of use DJ-MC does not require separate ratings for songs and
transitions.  Rather, it can assign credit to each component individually
from a single unified reward signal. It does so by computing the
relative contributions of the song and transition rewards to the total
reward as predicted by its model. This update procedure is presented in Algorithm 3.

Specifically, let $r$ be the reward the user assigns after hearing
song $a$ in state $s$, and $\bar{r}$ be the average rewards
assigned by this listener so far (line 4).  We define $r_{incr} =
log(\frac{r}{\bar{r}})$ (line 5). This factor determines both direction and
magnitude for the update (negative if $r < \bar{r}$, positive
otherwise, and greater the farther $r$ is from average). Let $R_s(a_i)$ and $R_t(a_{i-1}, a_i)$ be
 the expected song and transition rewards
yielded by our model, respectively. DJ-MC uses the proportions of these values to
set weights for credit assignment (this is essentially a maximum
likelihood estimate). Concretely, we define the update weights for the song and transition to be 

$ w_s = \frac{R_s(a_i)}{R_s(a_i) + R_t(a_{i-1},a_i)}$ and 

$ w_t = \frac{R_t(a_{i-1},a_i)}{R_s(a_i) + R_t(a_{i-1},a_i)}$ respectively (lines 6-7).

Finally, the agent uses the credit assignment
values determined at the previous step to partition the given reward
between song and transition weights, and update their
values (lines 8-9). Following this step, DJ-MC normalizes both the song and
transition reward models so that the weights for each feature sum up
to $1$ (line 10). This update procedure as a whole can be perceived as a temporal-difference
update with an attenuating learning rate, which balances how much the
model ``trusts'' the previous history of observations compared to the
newly obtained signal. It also guarantees convergence over time.

\begin{algorithm}[tb!]
   \caption{Model Update}
\label{alg:oneshot}

\begin{algorithmic}[1]
\STATE {\bfseries Input:} Song corpus, $\cal{M}$
 \vskip 1pt Planned playlist duration,  $K$
 \vskip 1pt
\FOR{ $i \in \{1, \ldots, K\}$}
\STATE Use Algorithm 4 to select song $a_i$, obtaining reward $r_i$
\STATE let $\bar{r} = average(\{r_1, \ldots, r_{i-1}\})$
\STATE $r_{incr}$ = $log(r_i/\bar{r})$
 \vskip 1pt weight update:
\STATE $ w_s = \frac{R_s(a_i)}{R_s(a_i) + R_t(a_{i-1},a_i)}$ 

\STATE $ w_t = \frac{R_t(a_{i-1},a_i)}{R_s(a_i) + R_t(a_{i-1},a_i)}$

\STATE $\phi_s = \frac{i}{i+1} \cdot \phi_s + \frac{1}{i+1} \cdot \theta_s \cdot w_s \cdot r_{incr}$

\STATE $\phi_t = \frac{i}{i+1} \cdot \phi_t + \frac{1}{i+1} \cdot \theta_t \cdot w_t \cdot r_{incr}$

\STATE Per $d \in \mbox{descriptors}$, normalize $\phi_s^d, \phi_t^d$ 
 \vskip 1pt (where $\phi_x^d$ denotes coordinates in $\phi_x$ corresponding to 10-percentile bins of descriptor $d$)
\ENDFOR

\end{algorithmic}

\end{algorithm}

\subsection{Planning} 


Equipped with the listener's learned song and transition utility
functions $R_s$ and $R_t$, which determine the MDP reward function
$R(s,a)=R_s(a)+R_t(s,a)$, DJ-MC employs a tree-search heuristic for
planning, similar to that used in \cite{AAMAS13-urieli}.
As in the case of initializing the transition weights (Algorithm 2),
DJ-MC chooses a subset of $50$-percent of the songs in the database,
which, based on $R_s$, obtain the highest song reward (line 2). At each point,
it simulates a trajectory of future songs selected at random from this
``high-yield'' subset (lines 7-11). The DJ-MC architecture then uses $R_s$ and $R_t$ to
calculate the expected payoff of the song trajectory (line 12). It repeats this
process as many times as possible, finding the randomly generated
trajectory which yields the highest expected payoff (lines 13-16). DJ-MC then
selects the first song of this trajectory to be the next song
played (line 19). It uses just the first song and not the whole sequence because
as modeling noise accumulates, its estimates become farther off. Furthermore, as we
discussed in Subsection $5.3$, DJ-MC actively adjusts $\phi_s$ and $\phi_t$ online based on user feedback using Algorithm 3. As a result, replanning at every step is advisable.

If the song space is too large or the search time is limited, it may
be infeasible to sample trajectories starting with all possible songs.
To mitigate this problem, DJ-MC exploits the structure of the song
space by clustering songs according to song types (line 9).\footnote{ In
principle, any clustering algorithm could work.  For our experiments,
we use the canonical k-means algorithm~\cite{macqueen1967some}.}  It
then plans over abstract song types rather than concrete songs, thus
drastically reducing search complexity. Once finding a promising
trajectory, DJ-MC selects a concrete representative from the first
song type in the trajectory to play (line 18).

\begin{algorithm}[tb!]
   \caption{Plan via Tree Search}
\label{alg:oneshot}

\begin{algorithmic}[1]
\STATE {\bfseries Input:} Song corpus $\cal{M}$, planning horizon $q$
\STATE Select upper median of $\cal{M}$, $\cal{M}^*$, based on $R_s$
\STATE \emph{BestTrajectory} $= null$
\STATE \emph{HighestExpectedPayoff} $= -\infty$
\WHILE {computational power not exhausted}
        \STATE {$trajectory = []$}
        \FOR {$1..\ldots q$}
                \STATE $song \leftarrow$ selected randomly from $\cal{M}^*$ 
                \vskip 1pt \emph{(avoiding repetitions)}
                \STATE optional: 
                 \vskip 1pt $song\_type \leftarrow$ selected randomly from $song\_types(\cal{M}^*)$
                                \vskip 1pt \emph{(avoiding repetitions, $song\_types(\cdot)$ reduces the set to clusters)}
                \STATE add $song$ to $trajectory$
        \ENDFOR
        \STATE $\mbox{expectedPayoffForTrajectory} = R_s(song_1) + \sum\limits_{i=2}^{q} (R_t((song_1, \ldots, song_{i-1}), song_i) + R_s(song_i))$
        \IF{$\mbox{expectedPayoffForTrajectory} > \mbox{HighestExpectedPayoff}$}
        \STATE $\mbox{HighestExpectedPayoff} = \mbox{expectedPayoffForTrajectory}$
        \STATE \emph{BestTrajectory} $ = $ \emph{trajectory}
        \ENDIF
\ENDWHILE
\STATE optional: if planning over song types, replace \emph{BestTrajectory}$[0]$ with concrete song.
\STATE return \emph{BestTrajectory}$[0]$
\end{algorithmic}
\end{algorithm}

Combining initialization, learning on the fly, and planning, the full DJ-MC agent architecture is presented in Algorithm 5.

\begin{algorithm}[tb!]
   \caption{Full DJ-MC Architecture}
\label{alg:oneshot}

\begin{algorithmic}[1]
\STATE {\bfseries Input:} $\cal{M}$ - song corpus, $K$ - planned playlist duration, $k_s$ - number of steps for song preference initialization, $k_t$ - the number of steps for transition preference initialization

\vskip \algorithmicindent Initialization:
\STATE Call Algorithm 1 with corpus $\cal{M}$ and parameter $k_s$ to initialize song weights $\phi_s$.
\STATE Call Algorithm 2 with corpus $\cal{M}$ and parameter $k_t$ to initialize transition weights $\phi_t$. 
\vskip\algorithmicindent Planning and Model Update:
\STATE Run Algorithm 3 with corpus $\cal{M}$ and parameter $K$
\vskip 1pt (Algorithm 3 iteratively selects the next song to play by calling algorithm 4, and then updates $R_s$ and $R_t$. This is repeated for $K$ steps.)
\end{algorithmic}
\end{algorithm}

\section{Evaluation in Simulation}

Due to the time and difficulty of human testing, especially in listening sessions lasting hours, it is important to first validate DJ-MC in simulation. To this end, we tested DJ-MC on a large set of listener models built using real playlists made by individuals and included in The Art of the Mix archive. For each experiment, we sample a $1000$-song corpus from the Million Song Dataset. 

One of the issues in analyzing the performance of DJ-MC was the nonexistence of suitable competing approaches to compare against. Possible alternatives are either commercial and proprietary, meaning their mechanics are unknown, or they do not fit the paradigm of online interaction with an unknown individual user. Still, we would like our evaluation to give convincing evidence that DJ-MC is capable of learning not only song preferences but also transition preferences to a reasonable degree, and that by taking transition into account DJ-MC is able to provide listeners with a significantly more enjoyable experience (see Section $8$ for related work).

In order to measure the improvement offered by our agent, we compare  DJ-MC  against two alternative baselines: an agent that chooses songs randomly, and a greedy agent that always plays the song with the highest song reward, as determined by Algorithm 1. As discussed in the introduction, we expect that the greedy agent will do quite well since song reward is the primary factor for listeners. However we find that by learning preferences over transitions,  DJ-MC yields a significant improvement over the greedy approach.

To represent different listener types, we generate $10$ different playlist clusters by using $k$-means clustering on the playlists (represented as artist frequency vectors). We generate $1000$ different listeners by first sampling a random cluster, second sampling $70\%$ of the song transition pairs in that cluster, and third inputting this data to Algorithms 1 and 2 to train the listener's song and transition weights. For the experiments reported here we used a playlist length of 30 songs, a planning horizon of 10 songs ahead, a computational budget of 100 random trajectories for planning, a query size of 10 songs for song reward modeling and 10 songs for transition rewards. As shown in Figure \emph{3},   DJ-MC  performs significantly better than the baselines, most noticeably in the beginning of the session.

\begin{figure}[h!]
  \label{fig1}
  \centering
    \includegraphics[height=200pt, width=0.425\textwidth,natwidth=310,natheight=322]{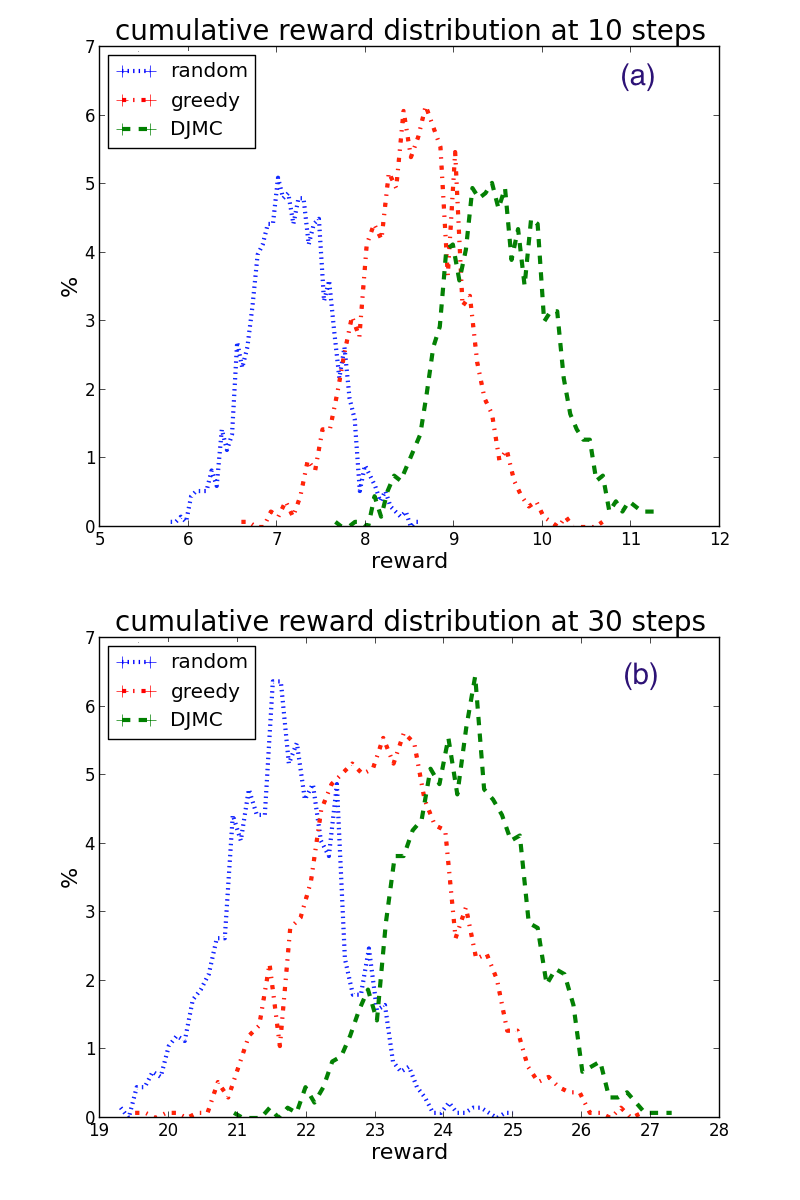}
      \caption{{\scriptsize Cumulative reward histograms for playlists of length $10$ (\emph{a}) and $30$ (\emph{b}), with listeners based on real playlist data. The DJ-MC agent outperforms both random and greedy agents, particularly for the first $10$ songs. Results are highly significant (\emph{p}-value $<<0.01$).}}
\end{figure}

\vspace{-0.5cm}
\section{Evaluation on Human Listeners}
While using simulated listeners allows for extensive analysis, ultimately the true measure of DJ-MC is whether it succeeds when applied on real listeners. To test whether this is the case, we ran two rounds of lab experiments with \emph{47} human participants. The participants pool was comprised of graduate students at the McCombs School of Business at the University of Texas at Austin.

\subsection{Experimental Setup}

Each participant interacted with a playlist generator. As a song corpus we used songs with corresponding Million Song Dataset entries that also appeared in Rolling Stone Magazine's list of 500 greatest albums of all time.\footnote{\url{http://www.rollingstone.com/music/lists/500-greatest-albums-of-all-time-20120531}} To keep the duration of the experiment reasonable, each song played for \emph{60} seconds before transitioning (with a cross-fade) to the next song. After each song the participants were asked, via a graphic user interface, to specify whether they liked or disliked the played song, as well as the transition to it. This provided us with separate (albeit not independent) signals for song quality and song transition quality to test how well DJ-MC actually did. Since asking users for their selection of $10$ songs was impractical in this setting, in order to seed the learning the agent explored randomly for \emph{25} songs, and then began exploiting the learned model (while continuing to learn) for \emph{25} songs. The participants were divided into \emph{2} groups - \emph{24} interacted with the greedy baseline, whereas \emph{23} interacted with DJ-MC. Though we expect the greedy agent to perform well based on song preferences only, we test whether DJ-MC's attention to transition preferences improves performance.

\subsection{Results}

Since our sample of human participants is not large, and given the extremely noisy nature of the input signals, and the complexity of the learning problem, it should come as no surprise that a straightforward analysis of the results can be difficult and inconclusive. To overcome this issue, we take advantage of bootstrap resampling, which is a highly common tool in the statistics literature to estimate underlying distributions using small samples and perform significance tests. 

At each stage we treat a ``like'' signal for either the transition or the song as $+1$ reward value vs. $0$ for a ``dislike''. We continue to reconstruct an approximate distribution of the aggregate reward for each condition by sampling subsets of \emph{8} participants with repetition for $N=250000$ times and estimating the average reward value for the subset. Figures \emph{4a} and \emph{4c} compare the reward distributions for the greedy and DJ-MC agents from song reward and transition reward respectively, during the first 25 episodes.  Since both act identically (randomly) during those episodes, the distributions are very close (and indeed testing the hypothesis that the two distributions have means more than $0.25$ apart by offsetting the distributions and running an appropriate t-test does not show significance). 

During the exploitation stage (episodes 26-50), the agents behave differently.  With regards to song reward, we see that both algorithms are again comparable (and better in expectation than in the exploration stage, implying some knowledge of preference has been learned), as seen in Figure \emph{4b}. In Figure \emph{4d}, however, we see that DJ-MC significantly outperforms the greedy algorithm in terms of transition reward, as expected, since the greedy algorithm does not learn transition preferences. The results are statistically significant using an unpaired t-test ($p << 0.01$), and are also significant when testing to see if the difference is greater than $0.25$. 

Interestingly, the average transition reward is higher for the greedy algorithm at the exploitation stage (apparent by higher average reward comparing Figures \emph{4a} and \emph{4b}). From this result we can deduce that either people are more likely to enjoy a transition if they enjoy the song, or that focusing on given tastes immediately reduces the ``risk'' of poor transitions by limiting the song space. All in all, these findings, made with the interaction of human listeners, corroborate our findings based on simulation, that reasoning about transition preferences gives DJ-MC a small but significant boost in performance compared to only reasoning about song preferences.

\begin{figure}[h!]
  \label{fig1}
  \centering
    \includegraphics[height=490pt,width=0.45\textwidth,natwidth=310,natheight=322]{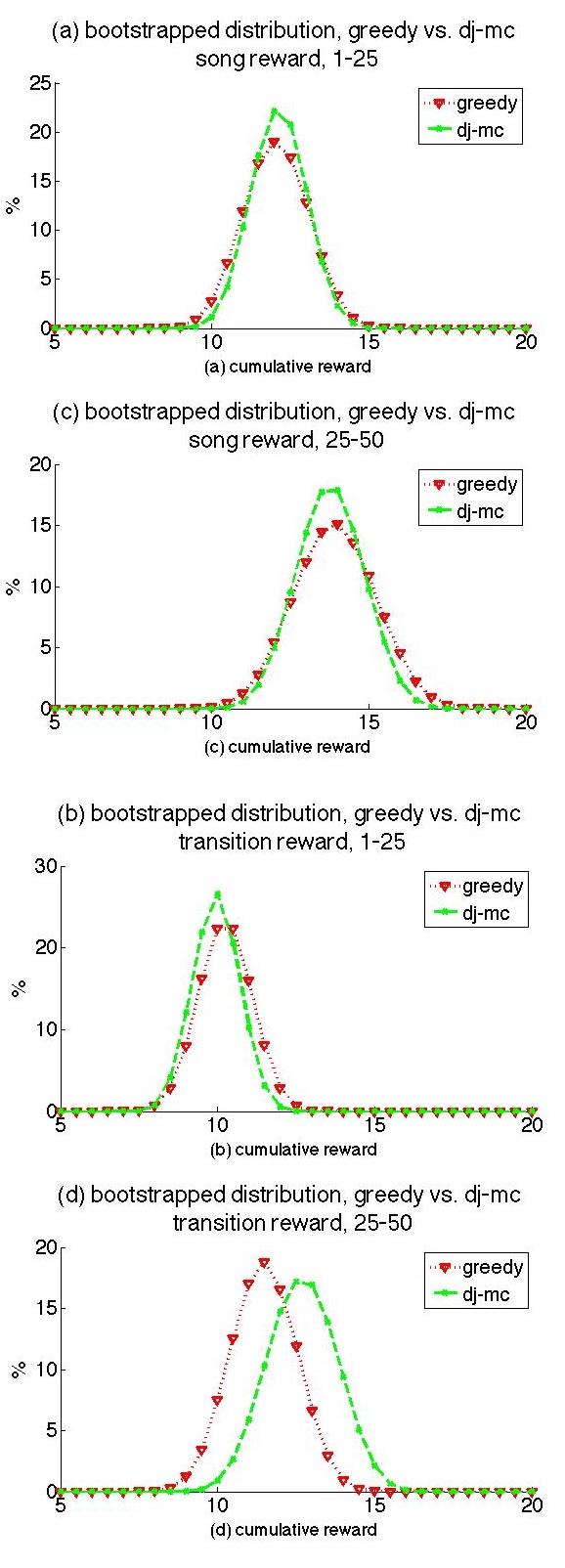}
      \caption{{\scriptsize (a) Histogram of cumulative song rewards for the first 25 songs (b) Histogram of cumulative song rewards for the songs 25-50. (c) Histogram of cumulative transition rewards for the first 25 songs. (d) Histogram of cumulative transition rewards for the songs 25-50. Histograms computed via bootstrap resampling the original data 250,000 times.}}
\end{figure}

\section{Related Work}

Though not much work has attempted to model playlists directly, there has been substantial research on modeling similarity between artists and between songs. Platt et al. \cite{platt2003fast} use semantic tags to learn a Gaussian process kernel function between pairs of songs. Weston et al. \cite{weston2011multi} learn an embedding in a shared space of social tags, acoustic features and artist entities by optimizing an evaluation metric for various music retrieval tasks. Aizenberg et al. \cite{aizenberg2012build} model radio stations as probability distributions of items to be played, embedded in an inner-product space, using real playlist histories for training.

In the academic literature, several recent papers have tried to tackle the issue of playlist prediction. Maillet et al.\cite{maillet2009steerable} approach the playlist prediction problem from a supervised binary classification perspective, with pairs of songs in sequence as positive examples and random pairs as negative ones. Mcfee and Lanckriet \cite{mcfee2011natural} consider playlists as a natural language induced over songs, training a bigram model for transitions and observing playlists as Markov chains. Chen et al.\cite{chen2012playlist} take on a similar Markov approach, treating playlists as Markov chains in some latent space, and learn a metric representation (or multiple representations) for each song in that space, without relying on audio data. In somewhat related work, Zheleva et al.\cite{zheleva2010statistical} adapt a Latent Dirichlet Allocation model to capture music taste from listening activities across users, and identify both the groups of songs associated with the specific taste and the groups of listeners who share the same taste. In a more recent related work, Natarajan et al. \cite{natarajan2013app} generalize this approach to the problem of collaborative filtering for interactional context. Users are clustered based on a one-step transition probability between items, and then transition information is generalized across clusters. Another recent work by Wang et al. \cite{wang2013exploration} also borrows from the reinforcement learning literature, and considers the problem of song recommendations as a bandit problem. Applying this approach, the authors attempt to balance the tradeoff between exploration and exploitation in personalized song recommendation.

The key difference between these approaches and our methodology is that to the best of our knowledge, no one has attempted to model entire playlists \emph{adaptively}, while interacting with a human listener individually and learning his preferences over both individual songs and song transitions online. By explicitly modeling transitions and exploiting user reinforcement, our framework is able to learn preference models for playlists on the fly without any prior knowledge.

\section{Summary and Discussion}

In this work we present DJ-MC, a full DJ framework, meant to learn the preferences of an individual listener online, and generate suitable playlists adaptively. In the experimental sections we show that our approach offers significant improvement over a more standard approach, which only considers song rewards. DJ-MC, which focuses on the audio properties of songs, has the advantage of being able to generate pleasing playlists that are unexpected with respect to traditional classifications based on genre, period, etc. In future work, it would be of interest to combine intrinsic sonic features with varied sources of metadata (e.g. genre, period, tags, social data, artist co-occurrence rates, etc). It would also be of interest to test our framework on specific types of listeners and music corpora. This work shows promise for both creating better music recommendation systems, and demonstrating the effectiveness of a reinforcement-learning based approach in a new practical domain.

\begin{scriptsize}

\subsection*{Acknowledgements}

This work has taken place in the Learning Agents Research
Group (LARG) at the Artificial Intelligence Laboratory, The University
of Texas at Austin.  LARG research is supported in part by grants from
the National Science Foundation (CNS-1330072, CNS-1305287), ONR
(21C184-01), AFOSR (FA8750-14-1-0070, FA9550-14-1-0087), and Yujin Robot.
\end{scriptsize}

\label{sec:disc}
\bibliographystyle{abbrv}

\bibliography{refs}

\end{document}